\documentclass[10pt,twocolumn,letterpaper]{article}

\usepackage[pagenumbers]{cvpr} 

\usepackage{graphicx}
\usepackage{amsmath}
\usepackage{amssymb}
\usepackage{booktabs}

\usepackage{algorithm}
\usepackage{listings}
\usepackage{etoolbox}
\usepackage[accsupp]{axessibility} 

\usepackage{xcolor}
\usepackage{enumitem}
\usepackage{xspace}
\usepackage{booktabs}
\usepackage{colortbl}
\usepackage{multirow}
\usepackage{makecell}
\usepackage{lipsum} 
\usepackage{gensymb} 

\usepackage[labelsep=period]{caption}
\renewcommand{\paragraph}[1]{\vspace{0.2em}\noindent \textbf{#1 \hspace{0.2em}}}

\definecolor{MyDarkRed}{rgb}{0.66, 0.16, 0.16}
\definecolor{MyDarkBlue}{rgb}{0.16, 0.16, 0.66}

\definecolor{cvprblue}{rgb}{0.21,0.49,0.74}

\usepackage[pagebackref=true,breaklinks=true,colorlinks=true,bookmarks=false,linkcolor={red!50!black},urlcolor={darkgray!80!black},citecolor={green!50!black}]{hyperref} 

\usepackage[marginal]{footmisc}

\newcommand{\wlink}[1]{\textcolor{magenta}{{#1}}}

\usepackage[capitalize]{cleveref}
\crefname{section}{Sec.}{Secs.}
\Crefname{section}{Section}{Sections}
\Crefname{table}{Table}{Tables}
\crefname{table}{Tab.}{Tabs.}

\def\paperID{3582} 
\def\confName{CVPR}
\def\confYear{2024}
\def\methodName{IPoD}

\title{{\methodName}: Implicit Field Learning with Point Diffusion for\\Generalizable 3D Object Reconstruction from Single RGB-D Images}

\author{Yushuang Wu$^{1,2}\footnotemark[1]$ \quad Luyue Shi$^{1,2}$ \quad Junhao Cai$^{4}$ \quad Weihao Yuan$^{3}$ \quad Lingteng Qiu$^{1,2}$ \\ Zilong Dong$^{3}$ \quad Liefeng Bo$^{3}$ \quad Shuguang Cui$^{2,1}$ \quad Xiaoguang Han$^{2,1}\footnotemark[2]$ \vspace{0.3em} \\
{\normalsize $^1$SSE, CUHKSZ}
\quad{\normalsize $^2$FNii, CUHKSZ} \quad {\normalsize $^3$Alibaba Group} \quad{\normalsize $^4$HKUST}
}

\begin{document}


\makeatletter
\AfterEndEnvironment{algorithm}{\let\@algcomment\relax}
\AtEndEnvironment{algorithm}{\kern2pt\hrule\relax\vskip3pt\@algcomment}
\let\@algcomment\relax
\newcommand\algcomment[1]{\def\@algcomment{\footnotesize#1}}
\renewcommand\fs@ruled{\def\@fs@cfont{\bfseries}\let\@fs@capt\floatc@ruled
  \def\@fs@pre{\hrule height.8pt depth0pt \kern2pt}%
  \def\@fs@post{}%
  \def\@fs@mid{\kern2pt\hrule\kern2pt}%
  \let\@fs@iftopcapt\iftrue}
\makeatother

\maketitle

\footnotetext[1]{Work done during internship supervised by Weihao Yuan at Alibaba.}
\footnotetext[2]{Corresponding author:
\wlink{hanxiaoguang@cuhk.edu.cn}.}

\begin{abstract}

    Generalizable 3D object reconstruction from single-view RGB-D images remains a challenging task, particularly with real-world data. 
    Current state-of-the-art methods develop Transformer-based implicit field learning, necessitating an intensive learning paradigm that requires dense query-supervision uniformly sampled throughout the entire space. 
    We propose a novel approach, {\methodName}, which harmonizes implicit field learning with point diffusion. 
    This approach treats the query points for implicit field learning as a noisy point cloud for iterative denoising, allowing for their dynamic adaptation to the target object shape. Such adaptive query points harness diffusion learning's capability for coarse shape recovery and also enhances the implicit representation's ability to delineate finer details. Besides, an additional self-conditioning mechanism is designed to use implicit predictions as the guidance of diffusion learning, leading to a cooperative system. 
    Experiments conducted on the CO3D-v2 dataset affirm the superiority of {\methodName}, achieving 7.8\% improvement in F-score and 28.6\% in Chamfer distance over existing methods. The generalizability of {\methodName} is also demonstrated on the MVImgNet dataset. Our project page is at \wlink{https://yushuang-wu.github.io/IPoD}.

\end{abstract}


\section{Introduction}
\label{sec:intro}
3D reconstruction from a single-view image is a challenging problem that with widespread implications in fields such as robotics, autonomous driving, and AR/VR. Recent efforts have been directed towards developing a generalizable model for object reconstruction from real RGB-D data~\cite{wu2023mcc, lionar2023numcc}. It aims to learn a category-agnostic network to recover an accurate and complete shape from a single-view object RGB-D image but with only imperfect ground-truth (GT) supervision, which are usually point clouds reconstructed from multiple views inevitably containing noise and incompleteness as in real 3D datasets~\cite{reizenstein2021co3d}.

\begin{figure}[tb] \centering
    \includegraphics[width=0.48\textwidth]{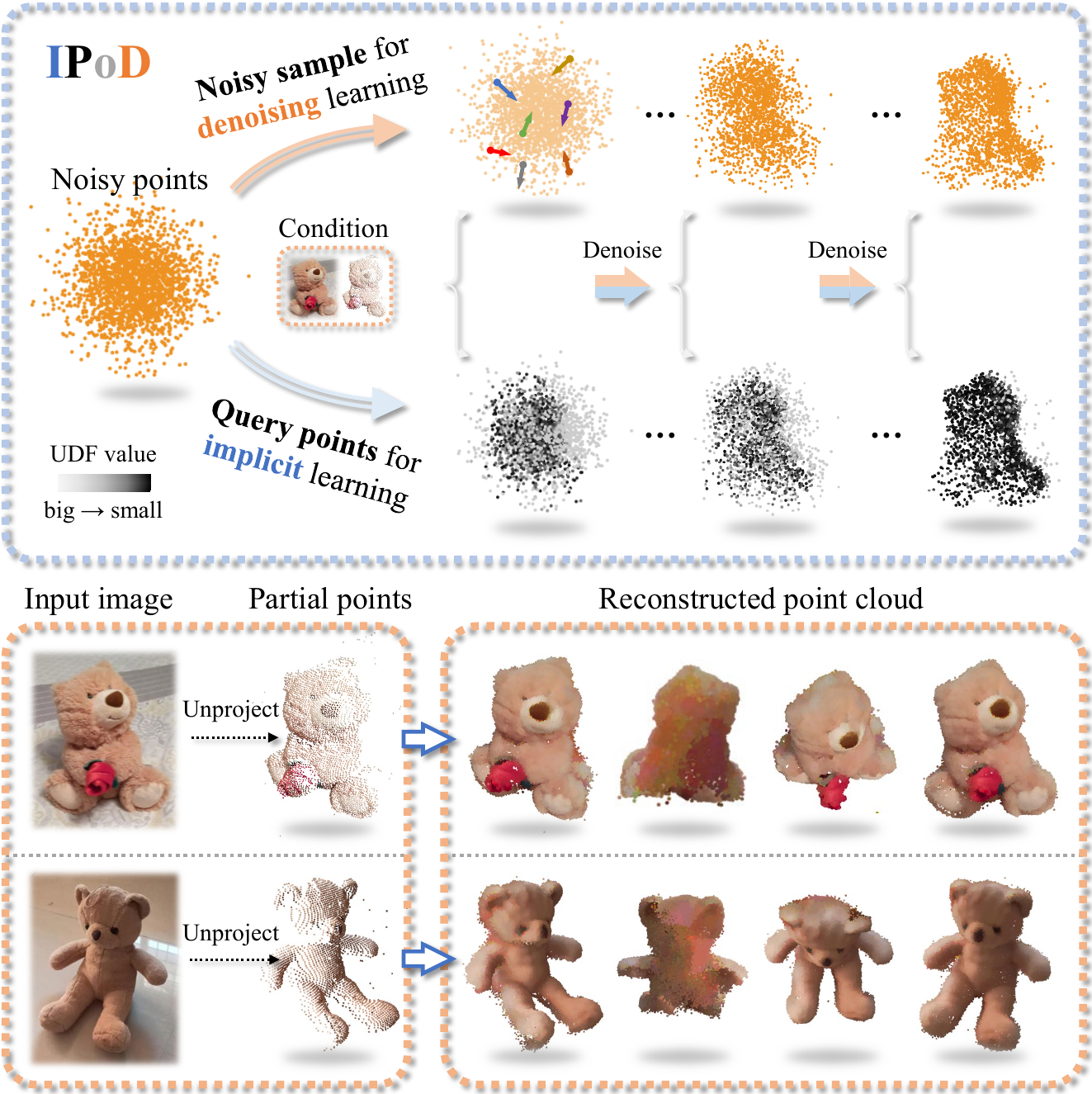}
    \caption{Our work focuses on the task of generalizable 3D object reconstruction from a single RGB-D image. The proposed method conducts implicit field learning with point diffusion that iteratively denoises a point cloud as adaptive query points for better implicit field learning, which leads to high reconstruction quality on both the global shape and fine details.} \label{fig:teaser}
    \vspace{-4mm}
\end{figure}

To tackle this problem, the state-of-the-art methods MCC~\cite{wu2023mcc} and NU-MCC~\cite{lionar2023numcc} develop Transformer-based networks to learn an implicit field for reconstruction. These methods involve an intensive learning paradigm that demands dense query-supervision sampled uniformly in a pre-defined bounded space. 
Simultaneously, denoising diffusion models, notable for their emergent role in generative modeling, have achieved remarkable outcomes in a variety of 2D and 3D tasks~\cite{dhariwal2021diffusion, ho2020ddpm, song2020ddim, saharia2022image, kawar2023imagic, liu2023zero, poole2022dreamfusion, lin2023magic3d}. The diffusion models prove powerful in generation especially given large amounts of data for training. Consequently, with the development of large-scale, realistic 3D datasets~\cite{reizenstein2021co3d, yu2023mvimgnet}, the integration of diffusion models into this field is anticipated to further enhance problem-solving capabilities. 

A subset of methods directly use 2D diffusion models for 3D reconstruction, which first pre-train an auto-encoder to represent object shapes into the latent space and learn to denosie a random noise of latent code (or triplane features) under specific conditions such as images or text~\cite{cheng2023sdfusion, chou2023diffusion, gupta20233dgen, shue2023triplane, zhou2023sparsefusion, chen2023ssdnerf, zeng2022lion}. The denoised latent code is subsequently utilized by the well-trained decoder to generate the reconstructed shape. Conversely, another approach simplifies this process by extending 2D diffusion into 3D~\cite{luo2021diffusion, zhou2021pvdiffusion, melas2023pc2, lyu2021diffrefine, zheng2023locallyatten}. It works on iteratively denoising a noisy 3D point cloud or voxels under given conditions to conduct reconstructing that has achieved remarkably generation results. Based on this, our work explores introducing the point diffusion into the target task. 

In this paper, we propose to integrate \underline{\textbf{I}}mplicit field learning with \underline{\textbf{Po}}int \underline{\textbf{D}}iffusion to address the target task, named as \textbf{{\methodName}}. The proposed integration approach is novel, simple, yet effective: we perceive the query points as a noisy point cloud to denoise. 
Specifically, a bunch of noisy points are sampled not only for denoising as in diffusion learning, but also as spatial queries for implicit field learning, so that our model iteratively conducts denoising and implicit predicting to recover the target shape under the RGB-D condition, as shown in Fig.~\ref{fig:teaser}. 
In this process, the sampled points can gradually get close to the true object shape through denoising, which can provide adaptive query positions according to the object shape for more effective implicit field learning. In comparison, a pure implicit field learning method queries all possible positions aimlessly and views them equally, while our method can more effectively conduct implicit field learning by attending more valuable local regions near the object surface, thus also capturing the fine shape details more easily. 
Further, we propose a novel self-conditioning mechanism~\cite{chen2023analog}, which leverages the predicted implicit values to reversely assist the diffusion learning and thus forges a cooperative system. 
Specifically, we predict the unsigned distances of all noisy points away from the true object surface and employ them as the condition of denoising at the next time step. The implicit predictions actually indicate the final target shape that is useful to guide every one-step denoising.  
The proposed method actually leads to a simple framework that conducts point diffusion learning and implicit field learning concurrently but well combines the advantages of both: the diffusion model for recovering the global coarse shape and the implicit field learning for giving accurate predictions on local fine details. 

We conduct experiments on the CO3D-v2~\cite{reizenstein2021co3d} dataset and demonstrate the superiority of the proposed approach, which surpasses the state-of-the-art results by $\sim$7.8\% of F-score and $\sim$28.6\% of Chamfer distance in average. 
In addition, we clean 100k point clouds reconstructed in MVImgNet~\cite{yu2023mvimgnet} and show that (i) trained on CO3D-v2, our method is generalizable to not only unseen categories in CO3D-v2 but also other various categories in MVImgNet; (ii) using the cleaned MVImgNet point clouds for training can help further improve the generalizability.

In summary, our key contributions are as follows:  
\begin{itemize}[itemsep=0pt,parsep=0pt,topsep=2bp]
    \item We propose {\methodName} that conducts implicit field learning with point diffusion for generalizable 3D object reconstruction from single RGB-D images, where the diffusion model provides adaptive queries for a more effective implicit field learning.
    \item We design a novel self-conditioning mechanism that leverages the implicit predictions to reversely assist the denoising thus leading to a mutually beneficial system. 
    \item We conduct extensive experiments to show the superiority of {\methodName} and the effectiveness of each component. {\methodName} achieves state-of-the-art reconstruction results on the CO3D-v2 dataset~\cite{reizenstein2021co3d}.
\end{itemize}

\section{Related Work}
\label{sec:related_works}

\paragraph{Single-view 3D reconstruction} 
One line of methods on this task train neural networks supervised by CAD~\cite{gkioxari2019meshrcnn, wang2018pixel2mesh}, voxels~\cite{girdhar2016learning, wu2017marrnet}, point clouds~\cite{fan2017point, mescheder2019occupancy}, or meshes~\cite{kulkarni2021s, xu2019disn}. However, they mainly focus on simplistic synthetic data with perfect supervision~\cite{wallace2019few, yan2016perspective}, \eg from the ShapeNet dataset~\cite{chang2015shapenet} or work on one or several categories~\cite{chen2019learning, liu2019soft, kanazawa2018learning, goel2020shape} as on the Pix3D dataset~\cite{sun2018pix3d}. Our work follows two more recent works, MCC~\cite{wu2023mcc} and NU-MCC~\cite{lionar2023numcc}, that tackle the problem of category-agnostic reconstruction from a single-view RGB-D image on real datasets~\cite{reizenstein2021co3d, yu2023mvimgnet}. MCC first introduces a large Transformer-based network for occupancy field learning, and NU-MCC further proposes a repulsive unsigned distance field (Rep-UDF) for finer reconstruction and a neighborhood decoder that speeds up the inference. Different in methodology, MCC-style methods take pure implicit learning for reconstruction, while our method integrates diffusion models for better reconstruction quality on both coarse shapes and fine details. 






\paragraph{Diffusion models} 
Since proposed in~\cite{ho2020ddpm, song2020ddim, nichol2021improvedddpm}, diffusion models have rapidly emerged as a popular family of generative models.
Methods based on diffusion technology have shown impressive quality, diversity, and expressiveness in various tasks such as image synthesis~\cite{dhariwal2021diffusion, ho2020ddpm, ho2022cascaded, nichol2021improvedddpm}, super-resolution~\cite{saharia2022image, li2022srdiff}, image editing~\cite{meng2021sdedit, sinha2021d2c, cheng2023adaptively, rombach2022high}, and 3D vision tasks like shape completion~\cite{lyu2021diffrefine, zhou2021pvdiffusion, chu2023diffcomplete}, text/image-to-3d~\cite{nam2022ldm3d, zheng2023locallyatten, poole2022dreamfusion, chen2023fantasia3d, gupta20233dgen}, and 3D reconstruction~\cite{cheng2023sdfusion, chou2023diffusion, shim2023sdfdiffusion, melas2023pc2, karnewar2023holodiffusion}.
Although applied on the 3D tasks, most methods develop 1D or 2D diffusion models by first pre-training an auto-encoder to encode the input into the 1D or 2D latent space and conduct diffusion learning on the latent space~\cite{cheng2023sdfusion, chou2023diffusion, gupta20233dgen, shue2023triplane, zhou2023sparsefusion, chen2023ssdnerf, zeng2022lion}. Zhou \textit{et al.}~\cite{zhou2021pvdiffusion} first extend diffusion models into 3D and propose point-voxel diffusion (PVD) to iteratively denoise a noisy point cloud for 3D shape generation and reconstruction. The 3D diffusion model is attracting increasing attention~\cite{luo2021diffusion, zhou2021pvdiffusion, melas2023pc2, lyu2021diffrefine, zheng2023locallyatten}. Closed to our target task, PC$^2$~\cite{melas2023pc2} applies PVD on real-world reconstruction from one image with a known camera pose, which projects image features to the point cloud as the diffusion condition. Different from our work, PC$^2$ focuses on reconstruction from posed images without depth and conducts category-specific learning. In methodology, our method is also based on point cloud diffusion models, but differently, ours integrates implicit predictions as the denoising guidance for more accurate noise estimation.

\paragraph{Implicit field learning} 
Implicit field learning is a powerful approach that learns an implicit field that assigns a value to each position as a representation. It has been widely used for 3D shape representation using occupancy field~\cite{peng2020convocc, niemeyer2019occupancy, saito2019pifu, wu2023mcc}, signed/unsigned distance field (SDF/UDF)~\cite{park2019deepsdf, chibane2020udf, lionar2023numcc}, implicit feature field~\cite{chibane2020implicit}, radiance field~\cite{mildenhall2020_nerf_eccv20, martin2021nerfwild}, etc. Although impressive results have been achieved, implicit field learning suffers from the large training cost because it usually demands dense position querying and implicit value supervising to well fit the field. Such an intensive learning paradigm is also used in the SOTA methods~\cite{wu2023mcc, lionar2023numcc} on the target task. However, our method integrates implicit field learning with diffusion learning, which can provide adaptive query positions via iteratively denoising the query points. In this way, the implicit field learning can more effectively extract the valuable information in data, which especially eases the recovery of fine details. 




\begin{figure*}[tb] \centering
    \includegraphics[width=\textwidth]{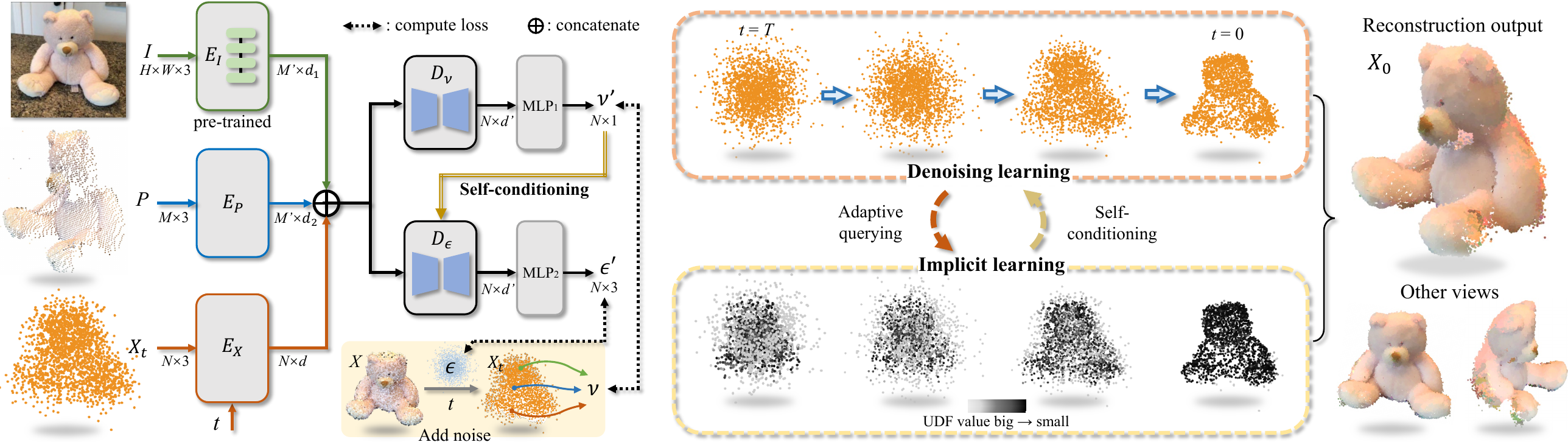}
    \caption{Overview of the proposed method. The network takes a single-view image and a partial point cloud unprojected from the image according to the depth information as the input. A bunch of points are sampled both as a noisy point cloud for point diffusion learning and also as query points for implicit field learning. The proposed self-conditioning mechanism leverages the implicit predictions to reversely assist denoising. The reconstruction result can be obtained via iteratively conducting implicit predicting and denoising concurrently.} \label{fig:overview}
    \vspace{-4mm}
\end{figure*}

\section{Method}%
\label{sec:method}
In this section, we first formulate the target problem and introduce the solutions based on implicit field learning and point diffusion models. Then we give the formulation of the proposed {\methodName} that integrates the two solutions. Finally, we introduce the design of our self-conditioning mechanism.

\subsection{Preliminary}%
\label{sub:preliminary}
\paragraph{Problem Formulation} The task of this work aims to recover a 3D point cloud $X \in \mathbb{R}^{N \times 3}$ from a RGB-D input, which is usually processed into an image $I \in [0, 255]^{H \times W \times 3}$ of size $H$ and $W$ and a partial point cloud $P \in \mathbb{R}^{M \times 3}$ unprojected from $I$ with depth information, where $N$ and $M$ denote the point numbers and $M=HW$ if without filtering or down-sampling. 
All point clouds are normalized with zero-mean and unit-variance as the CO3D~\cite{reizenstein2021co3d} coordinate system. 

\paragraph{Implicit Field Learning} This solution aims to learn an implicit field:
$f:(x,y,z) \rightarrow v$,
where $(x,y,z)$ denotes any position in the 3D space, $v$ is an implicit value, \eg a UDF value $udf \in \mathbb{R}^+$ representing the distance to the object surface. 
In training, a bunch of $N$ query points $Q = \{(x,y,z)\} \subseteq \mathrm{U}(-b, b)^3$ are uniformly sampled from a bounded space of scale $2b$, and their corresponding implicit values $\nu \in \mathbb{R}^{N\times 1}$ are computed as supervision. Thus given $P$ and $I$ as references, the implicit field learning network $f_\theta$ aims to learn:
\begin{align}
f_\theta(Q ~|~ P,I) \rightarrow \nu,
\label{equ:implicit}
\end{align}
so that the target shape can be obtained via densely sampling query points and preserving those with desired implicit values to derive an output point cloud. The objective function for training is usually to minimize an $L1$ distance:
\begin{align}
\mathcal{L}_\mathrm{imp} = \big \| f_\theta(Q ~|~ P,I) - \nu \big \|_1,
\label{loss:implicit}
\end{align}

\paragraph{Point Diffusion Models} Diffusion denoising probabilistic models are inspired by a thermodynamic diffusion process. It works by iteratively adding noise to a sample $X_0 \sim q(X_0)$ from the target data distribution $q(X_0)$ and finally into purely random noise, and the generative model is formed by reversing the Markovian noising process. The noising stepsize in the diffusion process is defined by a variance schedule $\{\beta_t\}_{t=0}^T$:
\begin{align}
q(X_t|X_{t-1}) = \mathcal{N}(X_t; \sqrt{1-\beta_t}X_{t-1},\beta_t \mathbf{I}),
\end{align}
where $q(X_t|X_{t-1})$ is a normal distribution, so that $q(X_t|X_0)$ can be modeled by a reparameterization trick:
\begin{align}
q(X_t|X_0) = \sqrt{\bar\alpha_t}X_0 + \epsilon \sqrt{1-\bar\alpha_t},
\label{equ:xt}
\end{align}
where $\alpha_t = 1-\beta_t$, $\bar\alpha_t = 1-\prod_{s=0}^t \alpha_s$, and the noise $\epsilon \sim \mathcal{N}(0, \mathbf{I})$.
A generative network $g_\theta$ is learned to approximate $q(X_{t-1}|X_t)$, so that a sample $X_0 \sim q(X_0)$ can be generated by starting from a random sample $X_T \sim \mathcal{N}(0, \mathbf{I})$ and then iteratively sampling from the estimated $q(X_{t-1}|X_t)$ for denoising. 

The target task can be tackled by a conditional point cloud diffusion model, where a network $g_\theta$ is learned to denoise a point cloud $X_T \sim \mathcal{N}(0, 1)^{N\times 3}$ sampled from a spherical Gaussian ball into the original object $X_0 = X$ given condition $P$ and $I$. At each denoising step $t$, $g_\theta$ is required to predict the noise $\epsilon \sim \mathcal{N}(0, 1)^{N \times 3}$ added in the most recent time step in $X_t$:
\begin{align}
g_\theta(X_t,t ~|~ P,I) \rightarrow \epsilon,
\label{equ:diffusion}
\end{align}
where both $X_t$ and $t$ act as the input of $g_\theta$. At the inference stage, we recover the mean value of the approximated $q(X_t|X_{t-1})$ from the prediction $g_\theta(X_t,t ~|~ P,I)$, and a sample from the distribution $q(X_t|X_{t-1})$ can be obtained with this mean to update $X_t$. When the time step gets sufficiently small, the denoised $X_t$ can well approximate the shape of $X$. The objective function for optimizing the parameters in a diffusion model $g_\theta$ is usually to minimize an $L2$ distance:
\begin{align}
\mathcal{L}_\mathrm{diff} = \big \| g_\theta(X_t,t ~|~ P,I) - \epsilon \big \|_2,
\label{loss:diffusion}
\end{align}

\subsection{Implicit Field Learning with Point Diffusion}%
We propose to integrate the implicit field learning and point diffusion into one framework, where a bunch of points are sampled not only as spatial queries for implicit field learning but also as a noisy sample for denoising to approximate the object shape. 

See Fig.~\ref{fig:overview} for an overview of the proposed {\methodName}. The spatial queries start from a randomly sampled noise that can be viewed as a point cloud $X_T \sim \mathcal{N}(0, 1)^{N\times 3}$. As in the inference stage of diffusion learning, it gradually gets close to the target shape via an iterative denoising process $t=T,T-1,\cdots,0$ under the condition of $I$ and $P$. A network conducts implicit value predicting on these adaptive queries $X_t$ concurrently and produces the final shape when $t=0$.
Denoting the new network as $h_\theta$, at any time step $t$, we estimate the noise $\epsilon$ in $X_t$ and the UDF value $\nu$ at the position of each point in $X_t$:
\begin{align}
h_\theta(X_t,t ~|~ P,I) \rightarrow (\epsilon, \nu).
\label{equ:combined}
\end{align}


For the training of $h_\theta$, we randomly sample a noise $\epsilon \sim \mathcal{N}(0, 1)^{N \times 3}$ and $t \in [1, T]$ at each iteration, and the noise $\epsilon$ is added into the GT point cloud $X$ to produce $X_t$ according to Eq.~\ref{equ:xt}. We also compute the distance of each point in $X_t$ to the nearest one in $X$ as the supervision $\nu$. Then the network $h_\theta$ takes $X_t$ and $t$ as input to estimate $\epsilon$ and $\nu$ conditioned by $P$ and $I$ as in Eq.~\ref{equ:combined}. We optimize the parameters in $h_\theta$ by jointly minimizing the losses in Eq.~\ref{loss:implicit} and Eq.~\ref{loss:diffusion}:
\begin{align}
\mathcal{L}_\mathrm{uni} = \big \| \nu' - \nu \big \|_1 + \lambda \big \| \epsilon' - \epsilon \big \|_2,
\label{loss:unified}
\end{align}
where $\nu'$, $\epsilon'$ are the prediction of $h_\theta$, and $\lambda \in \mathbb{R}^+$ is a weighting factor. Note that our model can also predict the target object color by extending the implicit value $\nu$ with RGB values $\nu = \{udf, rgb\} \in \mathbb{R}^{N\times 4}$, whose supervision can be obtained from the color of the nearest point in $X$ within a pre-defined small distance $\rho$. 

Compared with pure implicit field learning, ours exploits the generation power of point diffusion models to conduct more effective querying, where the query points can adaptively get close to the target shape rather than being sampled aimlessly. With the coarse shape indicated by the denoised query points, the network can better capture local fine details via implicit prediction.

\paragraph{Implementation} In the proposed framework, the concrete implementation of $h_\theta$ follows the work of \cite{melas2023pc2, wu2023mcc, lionar2023numcc}. We provide two versions of implementation based on PVCNN~\cite{liu2019pvcnn} and Transformer~\cite{vaswani2017attention}, respectively. As illustrated in Fig.~\ref{fig:imp_details}, the condition image $I$ is first fed into a Vision-Transformer~\cite{dosovitskiy2020vit} (ViT) encoder $E_I$ (well pre-trained and frozen), where a patch embedding is adopted to down-sample and serialize the image input and several Transformer layers then extract features in the shape of $M'\times d_1$, of which $M'$ is the sequence length and $d_1$ is the feature dimension. The encoding of $P$ is similar, an encoder $E_P$ extracts the features of $P$ in the shape of $M'\times d_2$. In the encoder $E_X$ for encoding $X_t$, the time step $t$ is first embedded into a vector of length $e'$ and a linear layer is adopted to project it into two values $\{scale, shift\}$, which then serve as the affine factors to transform the embedding of $X_t$. The final feature of $X_t$ is finally obtained through a linear layer. 
In the Transformer-based implementation, we employ the similar anchor prediction operation following NU-MCC~\cite{lionar2023numcc}, which further encode the features of $I$ and $P$ into $M''$ anchors with positions and features. Note that our method is independent to this operation. In the PVCNN-based implementation, we follow the projection manner in PC$^2$ to project the image features onto all positions in $P$ and $X_t$ and extend their features via concatenation into $(M+N) \times d$ with $d=d_1+ d_2$. In the decoding stage, we use two decoders with the same architecture except the input and output dimension for the UDF $\nu'$ and noise $\epsilon'$ prediction, respectively. The UDF prediction $\nu'$ is first computed and sent into the the other decoder, which is concatenated with the encoded feature of $X_t$ as the self-condition for the noise prediction. More details can be found in Sec.~\ref{sec:experiments} and supplementary materials.

\subsection{Self-conditioning}%
The adaptive queries provided by denoising benefit the implicit field learning to place greater importance on recovering the fine shape. We further propose a self-conditioning technique to exploit the implicit field learning to reciprocally assist the diffusion learning, thus forming a mutually beneficial system.  

The typical self-conditioning technique is first proposed in \cite{chen2023analog}, where the model is directly conditioned by the previously generated variable during the sampling process, \eg $\widetilde{X_0}$ namely the approximated $X_0$ at each time step as used in \cite{chen2023analog}. It is proposed to leverage the internal approximation to improve the prediction accuracy of the diffusion model.

We propose a novel self-conditioning method by taking the predicted implicit value $\nu'$ as the self-condition. Compared with using $\widetilde{X_0}$, ours can provide more accurate self-condition information, because the errors in $\widetilde{X_0}$ are usually significant when $t$ is large at both the training and inference stage, while the approximated implicit values $\nu$ are relatively independent of the time step variable. 

\begin{figure}[tb] \centering
    \includegraphics[width=0.48\textwidth]{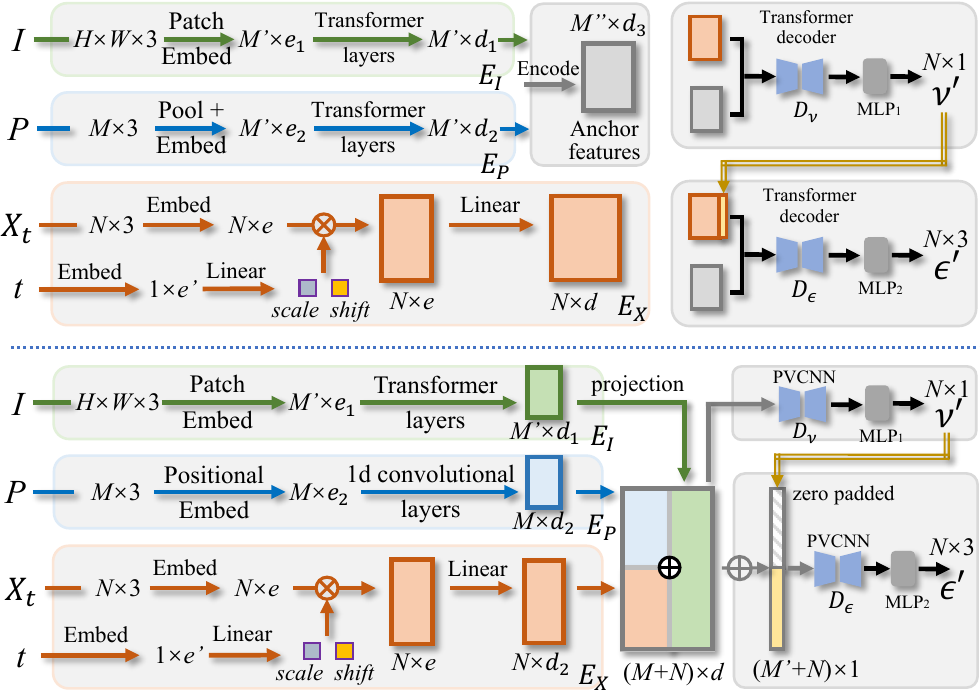}
    \caption{Illustration of the Transformer-based (upper part) and the PVCNN-based (lower part) implementations. $\otimes$ denotes the affine operation. The yellow arrow with double lines indicate the proposed self-conditioning mechanism. } \label{fig:imp_details}
    \vspace{-5mm}
\end{figure}

The self-conditioning mechanism works as illustrated in Fig.~\ref{fig:imp_details}. The UDF prediction $\nu'$ is first computed and sent into the the other decoder. As $\nu' \in \mathbb{R}^{N\times 1}$ provides point-wise information, we simply concatenate it with the embedding of $X_t$ in the feature dimension. The concatenated features then go through the decoder and an MLP for the noise prediction. At the inference stage, the self-condition is initialized with a vector with all negative values (\eg, -1) and updated with $\nu'$ at each time step. 

Compared with a naive denoising diffusion model as in Eq.~\ref{equ:diffusion}, the self-condition indicates the estimated unsigned distance of each point to the GT shape surface so that richer information about the target shape can be provided to assist the noise prediction.

\begin{table*}[tb]\centering
    \newcommand{\Frst}[1]{\textcolor{red}{\textbf{#1}}}
    \newcommand{\Scnd}[1]{\textcolor{blue}{\textbf{#1}}}
    \newcommand{\PTcnn}{~~PVCNN~~~}
    \newcommand{\Trans}{~~Transformer~~~}
    \caption{Results on CO3D-v2, averaged on all samples from 10 held-out categories. The best results are highlighted in bold font. }
    \label{tab:res_co3d}

\resizebox{0.8\textwidth}{!}{
    \Large
\begin{tabular}{l||c|*{3}{c}|*{3}{c}}
    \toprule
    Method & ~Backbone~ & ~~~~Acc$\downarrow$~~~~ & ~~~Comp$\downarrow$~~~ & ~~~~CD$\downarrow$~~~~ & ~~~~Prec$\uparrow$~~~~ & ~~~Recall$\uparrow$~~~ & ~~~~~F1$\uparrow$~~~~~ \\
    \midrule
    PC$^2$~\cite{melas2023pc2}         & \PTcnn & 0.342 & 0.214 & 0.556 & 24.2 & 56.2 & 33.0 \\
    PC$^2$-depth                       & \PTcnn & 0.209 & 0.103 & 0.312 & 61.7 & 87.6 & 70.7 \\
    MCC~\cite{wu2023mcc}               & \Trans & 0.172 & 0.144 & 0.316 & 68.9 & 72.7 & 69.8 \\
    NU-MCC~\cite{lionar2023numcc}~~~   & \Trans & 0.121 & 0.146 & 0.266 & 79.2 & 84.0 & 80.9 \\
    \midrule
    Ours1                              & \PTcnn & 0.163 & 0.089 & 0.252 & 69.0 & 89.7 & 76.2 \\
    Ours2                              & \Trans & \textbf{0.104} & \textbf{0.087} & \textbf{0.190} & \textbf{85.1} & \textbf{90.1} & \textbf{87.2} \\
    \bottomrule
\end{tabular}
}
    \vspace{-2mm}
\end{table*}

\section{Experiments}%
\label{sec:experiments}
\paragraph{Datasets} We use CO3D-v2~\cite{reizenstein2021co3d} as the main dataset following MCC~\cite{wu2023mcc} and NU-MCC~\cite{lionar2023numcc}. It consists of around 37k videos of 51 object categories, of which 10 are held out for evaluation and the remaining 41 for training. The 10 held-out categories are the same as the ones selected by MCC and NU-MCC, as listed in the supplementary materials.  In CO3D-v2, the object shape annotations are obtained via COLMAP~\cite{schonberger2016sfm, schonberger2016pixelwise} and thus inevitably contain noise and voids. We test the zero-shot generalization ability of the proposed method on the dataset of MVImgNet~\cite{yu2023mvimgnet}, which is a real-world dataset with 220k object videos in 238 categories, and their 3D annotations are also obtained via COLMAP. Note that videos of MVImgNet are captured in 180$\degree$ and thus can only provide incomplete 3D shape annotations. We also contribute a dataset with 100k cleaned point clouds from MVImgNet. We hire annotators to manually filter the 3D annotations with low quality and remove the background noise caused by COLMAP estimation for the rest of the point clouds.

\paragraph{Evaluation metrics} Following previous works~\cite{murez2020atlas, sun2021neuralrecon, wu2023mcc, lionar2023numcc}, we use 6 metrics to evaluate the reconstruction quality. The metrics can be divided into two groups for measuring (i) the absolute distance: the Chamfer distance (CD) and its two components that measure the distance in two different directions (Acc and Comp); and (ii) the relative degree of recovery: the precision ratio (Prec) that indicates the percentage of predicted points within a small distance threshold $\rho$ to any GT point, the recall ratio (Recall) that indicates the percentage of GT points within $\rho$ to any predicted point, and their F-score (F1). Among them, F1 holds the dominant since both accuracy and completeness are considered. The mathematical formulation of all metrics are detailed in the supplementary materials.

\paragraph{Baselines} We compare the proposed method with four baselines. The first one is PC$^2$~\cite{melas2023pc2} based on the diffusion model, implemented with a PVCNN backbone. It is designed for single-image reconstruction and conducts category-specific learning. We propose a modified PC$^2$ (PC$^2$-depth) by integrating the depth condition into the diffusion model, where the single-view depth is unprojected into a partial point cloud as the diffusion condition and concatenated with $X_t$ before being fed into the PVCNN for $X_t$ denoising. Another baseline is MCC~\cite{wu2023mcc} which is implemented with a Transformer-based encoder-decoder, which conducts implicit learning based on an occupancy field. The last one is NU-MCC~\cite{lionar2023numcc} that improves MCC by proposing a Repulsive UDF to replace the occupancy field and applying anchor representations for higher efficiency.

\paragraph{Implementation details} The size of input image $I$ is 224$\times$224, $P$ and $I$ are divided into patches of size 16$\times$16 for ViT encoding. The encoded feature dimensions of $I$ and $P$ are 768 ($d_1$=$d_2$=768), and the decoded feature dimension is also 768 ($d'$=768). The UDF value supervision is clamped with a max value of 0.5. Besides, $T$ is set as 1,000 in our diffusion model, $\lambda$=1.0 in training, and $N$=50k, the distance threshold $\rho$=0.1 for evaluation. Our model is trained with a batch size of 64 for 100 epochs (taking around 48 hours on NVIDIA V100 GPUs), and an Adam~\cite{kingma2014adam} optimizer with a base learning rate of 10$^{-4}$ is used following MCC and NU-MCC. We implement the proposed approach with two versions that use PVCNN (Ours1) and Transformer (Ours2) as the decoder backbone, respectively. The version of ``Ours1'' is constructed by adding the implicit learning branch and the self-conditioning mechanism into the diffusion-based PC$^2$~\cite{melas2023pc2}, and for ``Ours2'', we integrate the query points denoising branch into NU-MCC~\cite{lionar2023numcc} that constructed based on implicit field learning and use the predicted UDF values as the self-condition to guide the denoising. See supplementary materials for more model architecture details of PVCNN-based and Transformer-based implementations.

\begin{figure}[t] \centering
    \includegraphics[width=0.48\textwidth]{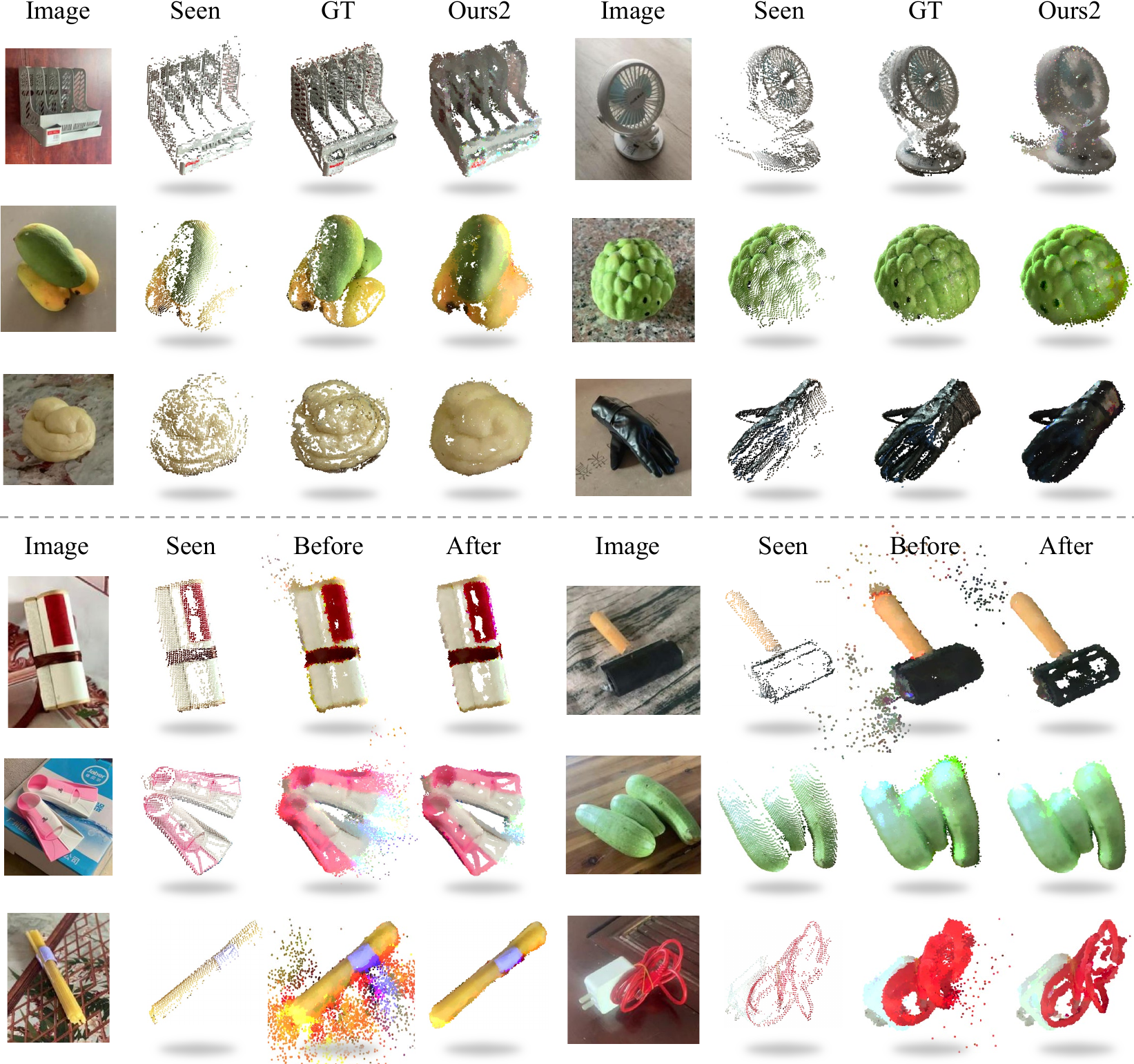}
    \caption{Visualization on MVImgNet data. Upper: generalization results by Ours2; Lower: the comparison of generalization results before and after fine-tuning on cleaned MVImgNet data.} 
    \label{fig:vis_mv}
    \vspace{-4mm}
\end{figure}

\subsection{Results on CO3D-v2}%
\label{sub:Results co3d}
We show the evaluation results of the proposed method on the CO3D-v2 dataset in Tab.~\ref{tab:res_co3d}. No matter when implemented with PVCNN or Transformer as the backbone, our method can get better results than other existing methods. With PVCNN, our method improves the performance of the baseline PC$^2$-depth by 19.2\% on Chamfer distance and 7.8\% on F-score. Based on Transformer, our method achieves SOTA performance, which surpasses the previously best algorithm NU-MCC overall metrics, specifically by 28.6\% on Chamfer distance (0.266$\rightarrow$0.190) and by 7.8\% on F-score (80.9\%$\rightarrow$87.2\%).

\begin{figure*}[tb] \centering
    \includegraphics[width=1.\textwidth]{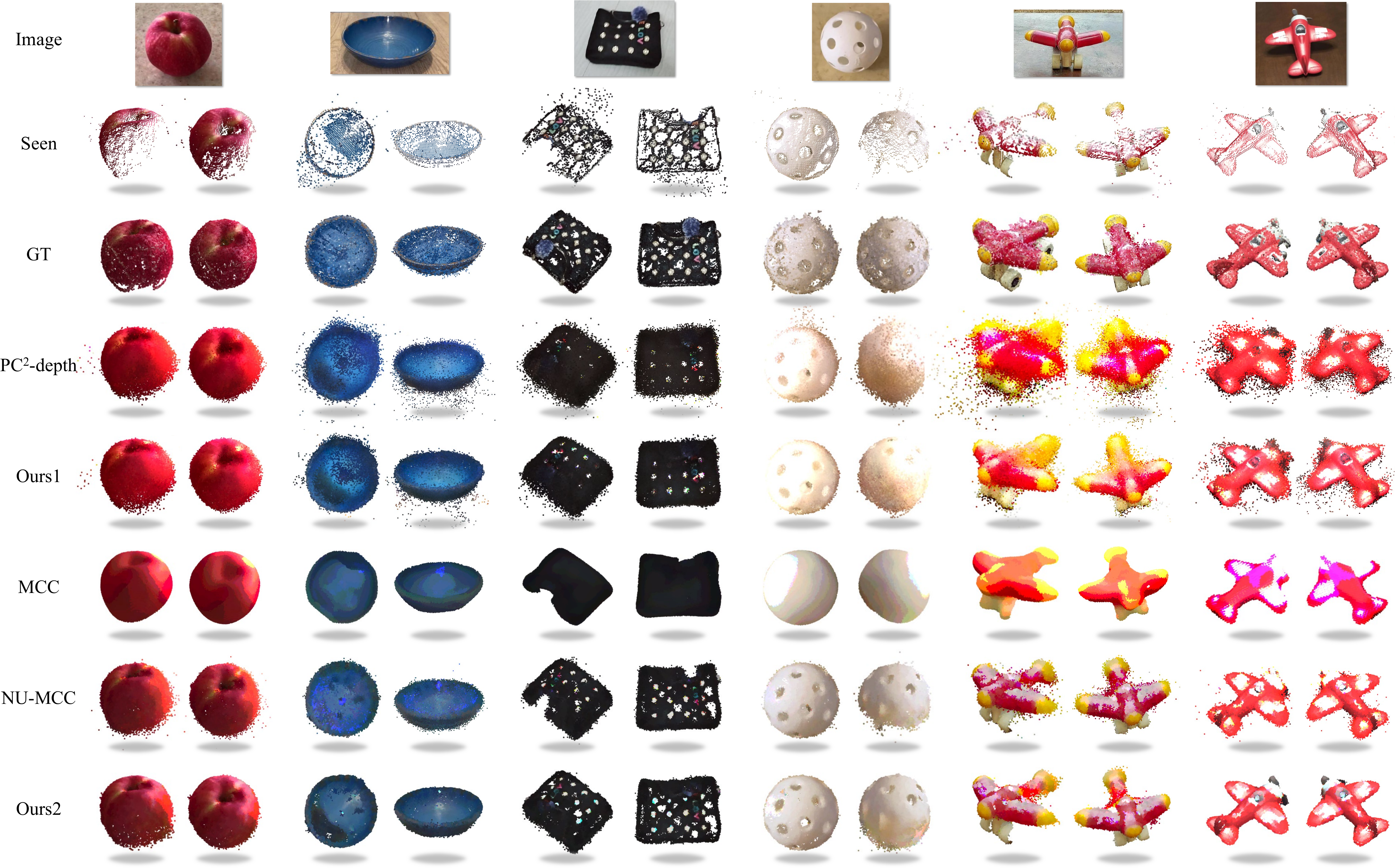}
    \caption{Visualization of reconstructions by different methods on CO3D-v2 unseen categories. We choose two views for each sample.} \label{fig:vis_compare}
    \vspace{-3mm}
\end{figure*}

We visualize the reconstruction results of each method in Fig.~\ref{fig:vis_compare}. As shown, the proposed method can produce better generations than the baseline method (Ours1 v.s. PC$^2$-depth and Ours2 v.s. NU-MCC) on both higher precision in local details and better completeness in coarse shapes. Besides, our method implemented based on Transformer achieve the best qualitative results over all baseline methods, especially for objects with more complex geometry structures. More qualitative results of our method on the CO3D-v2 dataset are included in the supplementary materials. We also visualize reconstructions of samples from seen categories in the supplementary materials.

\subsection{Results on MVImgNet}%
\label{sub:Results mv}
We evaluate the generalizability of our method implemented based on Transformer (Ours2) on the MVImgNet dataset. Considering the incompleteness of GT shapes in MVImgNet, we do not consider the quantitative results. As presented in Fig.~\ref{fig:vis_mv}, our method can well generalize to more various categories of objects than in the CO3D-v2 dataset. Meanwhile, there are also some cases of over-predicting as in the lower part of Fig.~\ref{fig:vis_mv}, where Ours2 predicts extra noisy points in results. We further use the cleaned MVImgNet point clouds to fine-tune the network and found that the generations are endowed with higher accuracy, which indicates that the cleaned data can further improve the model's generalizability. Note that the categories in the cleaned data for fine-tuning have no overlap with the ones for evaluation. This experiment can also validate the generalizability of point diffusion models when learning from large-scale data. More generalization results of our method are included in the supplementary materials. 

\begin{figure}[t] \centering
    \includegraphics[width=0.48\textwidth]{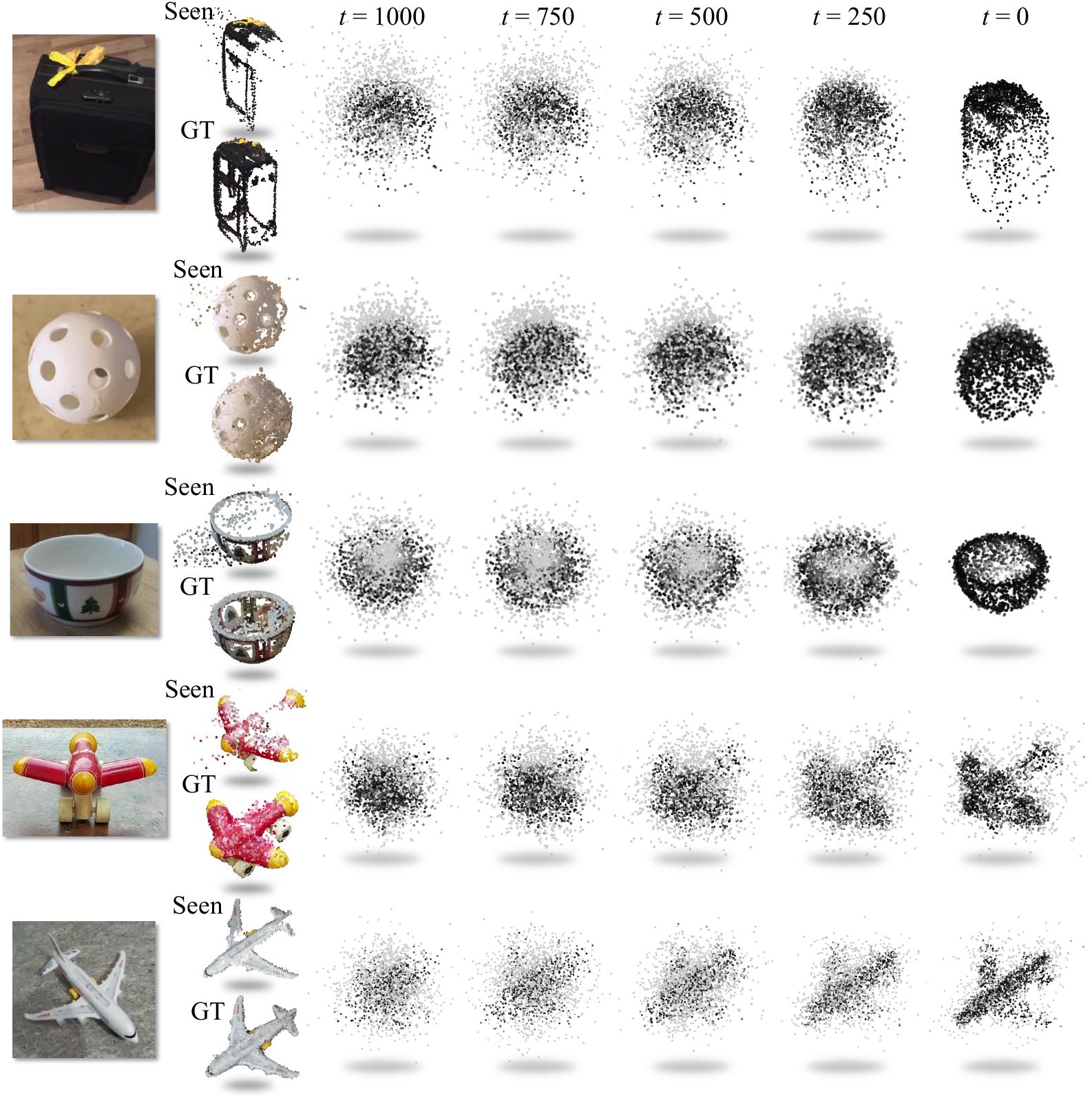}
    \caption{Visualization of the denoising process ($t$=$\{$1000, 750, 500, 250, 0$\}$) of our method in inferring. Note we only sample 2k points in each noisy sample for better visualization. The darkness of each point indicates the magnitude of the predicted UDF value (the smaller, the darker) at this position. } 
    \label{fig:vis_dp}
    \vspace{-4mm}
\end{figure}

\subsection{Ablation Study}%
Our approach mainly consists of three components: (i) the point diffusion learning provides adaptive query points for more efficient implicit learning, (ii) we supervise the UDF value of each point in a noisy point cloud for implicit field learning, and (iii) a self-conditioning mechanism is designed to facilitate its benefit to the denoising learning. Ablative analyses are conducted on them.

\paragraph{Individual impact} 
To analyze the impact of the three components above, we evaluate the precision, recall, and F-score of each variant. Adding diffusion learning only into NU-MCC can bring an obvious improvement by absolute 4.9\% on F-score (80.9\%$\rightarrow$85.8\%), and further adding self-conditioning also makes a positive effect (85.8\%$\rightarrow$87.2\%), as shown in Tab.~\ref{tab:abl_component}(c, d). When adding UDF learning into PC$^2$-depth, the F-score is raised absolutely by 2.6\% (70.7\%$\rightarrow$73.3\%), and adding self-conditioning can further bring an absolute 2.9\% gain (73.3\%$\rightarrow$76.2\%), which proves the effectiveness of self-conditioning. 

\begin{table}[tb]\centering
    \caption{Ablative results on three components: (i) Diff.: using diffusion learning; (ii) UDF: using UDF supervision; (iii) Self.: using the proposed self-conditioning mechanism. The upper section of the table includes results when implemented with PVCNN, and the lower includes Transformer-based results. }
    \label{tab:abl_component}
    \resizebox{0.48\textwidth}{!}{
    \Large
\begin{tabular}{c||*{3}{c}|*{3}{c}}
    \toprule
    Method~~~~~ & Diff. & ~UDF~ & Self. & ~~~Prec$\uparrow$~~~ & ~~Recall$\uparrow$~~ & ~~~~F1$\uparrow$~~~~ \\
    \midrule
    PC$^2$-depth& \checkmark &            &            & 61.7 & 87.6 & 70.7 \\
    (a)         & \checkmark & \checkmark &            & 65.8 & 87.4 & 73.3 \\
    (b)         & \checkmark & \checkmark & \checkmark & \textbf{69.0} & \textbf{89.7} & \textbf{76.2} \\
    \midrule
    NU-MCC~~    &            & \checkmark &            & 79.2 & 84.0 & 80.9 \\
    (c)         & \checkmark & \checkmark &            & 84.2 & 85.9 & 85.8 \\
    (d)         & \checkmark & \checkmark & \checkmark & \textbf{85.1} & \textbf{90.1} & \textbf{87.2} \\
    \bottomrule
\end{tabular}
}
    \vspace{-2mm}
\end{table}

\paragraph{Point diffusion} 
We visualize the denoising process in Fig.~\ref{fig:vis_dp} to better validate the effectiveness of the point diffusion learning. Starting from purely noisy samples, the point clouds gradually get close to the target shapes according to the inputs. Notably, when addressing objects with relatively simple geometry, the point cloud at $t=0$ itself can well capture the shape. However, for more complex geometries (\eg toy plane in the last two rows of Fig.~\ref{fig:vis_dp}), the noise in point clouds can not be perfectly diminished. In this situation, implicit values can still well indicate the accurate shapes that complement the denoised point clouds.

\begin{table}[tb]\centering
    \caption{Results of using different variants of self-condition. ``None'' denotes not using any self-conditioning.}
    \label{tab:abl_selfcond}
    \resizebox{0.46\textwidth}{!}{
    \Large
\begin{tabular}{l||*{3}{c}}
    \toprule
    Self-condition   &  ~~~~Prec$\uparrow$~~~~ & ~~~Recall$\uparrow$~~~ & ~~~~~F1$\uparrow$~~~~~ \\
    \midrule
    (a)~ UDF value                           & \textbf{69.0} & 89.7 & \textbf{76.2} \\
    (b)~ UDF~+~RGB value~~~                  & 68.9 & 89.8 & \textbf{76.2} \\
    (c)~ Occupancy value                     & 67.8 & 87.8 & 74.5 \\
    (d)~ $\widetilde{X_0}$ (classical)       & 63.2 & \textbf{90.3} & 72.7 \\
    (e)~ None                                & 61.7 & 87.6 & 70.7 \\
    \bottomrule
\end{tabular}
}
    \vspace{-2mm}
\end{table}

\begin{table}[tb]\centering
    \caption{Effects of self-conditioning at different denoising stages. The whole denoising process with 1k steps in inferring is evenly divided into four stages.}
    \label{tab:abl_timestep}
    \resizebox{0.48\textwidth}{!}{
    \Large
\begin{tabular}{*{4}{c}|*{3}{c}}
    \toprule
    1k-750 & 750-500 & 500-250 & ~250-0~ & ~~Prec$\uparrow$~~ & ~~Recall$\uparrow$~~ & ~~~F1$\uparrow$~~~ \\
    \midrule
               & \checkmark & \checkmark & \checkmark & 67.5 & 88.8 & 75.1 \\
    \checkmark &            & \checkmark & \checkmark & 56.0 & 86.7 & 65.5 \\
    \checkmark & \checkmark &            & \checkmark & 63.7 & 88.0 & 72.3 \\
    \checkmark & \checkmark & \checkmark &            & 59.9 & 85.1 & 68.1 \\
    \checkmark & \checkmark & \checkmark & \checkmark & \textbf{69.0} & \textbf{89.7} & \textbf{76.2} \\
    \bottomrule
\end{tabular}
}
    \vspace{-5mm}
\end{table}

\paragraph{Self-conditioning} 
We conduct another group of analysis on the choice of self-condition: (a) using predicted UDF values (ours), (b) using predicted UDF and RGB values, (c) using predicted Occupancy values, (d) using $\widetilde{X_0}$ as in a classical manner, and (e) using none of them above. As shown in Tab.~\ref{tab:abl_selfcond}, using RGB values additionally can not lead to an extra performance gain, and replacing UDF with the Occupancy field can cause a performance drop. A potential reason is that UDF values can provide more fine-grained information about the difference between the noisy sample and the target shape. Using $\widetilde{X_0}$ results in a worse F-score than using occupancy values, but better than not using. This result validates the claim in Sec.~\ref{sec:intro} that the proposed self-conditioning mechanism can provide more accurate and useful information about the target shape to improve the noise prediction in point diffusion learning.

Besides, we also delve into an in-depth analysis of the impact of self-conditioning at different denoising stages. Dividing the whole denoising process into four stages of even length (per 250 time steps), we close the self-conditioning at each single stage in inferring, which leads to the results in Tab.~\ref{tab:abl_timestep}. As shown, the first stage has the smallest effect, while the second and the last stages have significant impacts on the reconstruction quality. A possible reason is the second stage takes a role in capturing the coarse shape from a pure noise and the last stage is important for points' getting close enough to the target shape to ease the final UDF estimating.

\section{Conclusion}%
\label{sec:conclusion}
In this paper, we have introduced \methodName, a powerful approach based on implicit field learning and point diffusion, to address the problem of generalizable 3D object reconstruction from single RGB-D images. We propose to conduct implicit field learning with adaptive queries through point denoising that helps the model better capture both the global coarse shape and local fine details. We also develop a self-conditioning mechanism to leverage implicit predictions to reversely assist the noise estimation in diffusion learning, which eventually forges a cooperative system. We implement our method based on two kinds of backbones, PVCNN and Transformer, and evaluate them on the CO3D-v2 dataset. Experiments show that our method can achieve impressive reconstruction results, which quantitatively outperforms previous SOTA methods on both reconstruction completeness and precision. Results on MVImgNet also show that our method can well generalize to more various categories from another dataset.

\noindent
\textbf{Limitations} We have not validated the effectiveness of our method on 3D human and scene reconstruction. Human shapes often contain more fine-grained details that may bring new challenges, and 3D scenes are also hard to reconstruct considering the serve occlusion causing poor quality in single-view inputs. These will be our future work.

\paragraph{Acknowledgement} \small The work was supported in part by NSFC-62172348, the Basic Research Project No.\,HZQB-KCZYZ-2021067 of Hetao Shenzhen-HK S\&T Cooperation Zone, Guangdong Provincial Outstanding Youth Project No.~2023B1515020055, the National Key R\&D Program of China with grant No.\,2018YFB1800800, by Shenzhen Outstanding Talents Training Fund 202002, by Guangdong Research Projects No.\,2017ZT07X152 and No.\,2019CX01X104, by Key Area R\&D Program of Guangdong Province (Grant No.\,2018B030338001), by the Guangdong Provincial Key Laboratory of Future Networks of Intelligence (Grant No.\,2022B1212010001), and by Shenzhen Key Laboratory of Big Data and Artificial Intelligence (Grant No.\,ZDSYS201707251409055). It is also partly supported by NSFC-61931024
and Shenzhen General Project No.\,JCYJ20220530143604010.

{
    \small
    \bibliographystyle{ieeenat_fullname}
    \bibliography{ref}
}


\end{document}